\documentclass[conference]{IEEEtran}
\IEEEoverridecommandlockouts

\usepackage{cite}
\usepackage{comment}
\usepackage{multirow}

\usepackage{amsmath,amssymb,amsfonts}
\usepackage{algorithmic}
\usepackage{graphicx}
\usepackage{textcomp}
\usepackage{xcolor}
\usepackage{tikz}
\usepackage{hyperref}
\usepackage{soul}
\def\BibTeX{{\rm B\kern-.05em{\sc i\kern-.025em b}\kern-.08em
    T\kern-.1667em\lower.7ex\hbox{E}\kern-.125emX}}
\usepackage{pifont}%
\newcommand{\yes}{\ding{51}}%
\newcommand{\no}{\ding{55}}%
\newcommand{\mebots}{MEbots}	
\usepackage{color, colortbl}	
\definecolor{LightCyan}{rgb}{0.88,1,1}

\newcommand*\circled[1]{\tikz[baseline=(char.base)]{
            \node[shape=circle,draw,inner sep=1pt,fill={rgb,255:red,21;green,96;blue,130},text=white] (char) {#1};}}

\usepackage{tikz}
\usepackage{textcomp}
\usepackage[doipre={DOI:~}]{uri}
\usepackage{lipsum}
\newcommand\copyrighttext{%
  \footnotesize \textcopyright 2025 IEEE. Personal use of this material is permitted.  Permission from IEEE must be obtained for all other uses, in any current or future media, including reprinting/republishing this material for advertising or promotional purposes, creating new collective works, for resale or redistribution to servers or lists, or reuse of any copyrighted component of this work in other works.
 
  Accepted for publications at ISLPED 2025 IEEE/ACM International Symposium on Low Power Electronics and Design.}
\newcommand{\copyrightnotice}{%
\begin{tikzpicture}[remember picture,overlay]
\node[anchor=south,yshift=10pt] at (current page.south) {\fbox{\parbox{\dimexpr\textwidth-\fboxsep-\fboxrule\relax}{\copyrighttext}}};
\end{tikzpicture}%
}
\begin{document}
\bstctlcite{IEEEexample:BSTcontrol}
\title{
\mebots: Integrating a RISC-V Virtual Platform with a Robotic Simulator for Energy-aware Design
\thanks{This work has received funding from the Key Digital Technologies Joint Undertaking (KDT-JU) under grant agreement No 101095947 and grant agreement No 101112274. The JU receives support from the European Union’s Horizon Europe research and innovation program.

This publication is part of the project PNRR-NGEU which has received funding from the MUR – DM 117/2023.}
}
\author{\IEEEauthorblockN{
Giovanni Pollo\IEEEauthorrefmark{1}, Mohamed Amine Hamdi\IEEEauthorrefmark{1}, Matteo Risso\IEEEauthorrefmark{1}, Lorenzo Ruotolo\IEEEauthorrefmark{1}, Pietro Furbatto\IEEEauthorrefmark{1}, Matteo Isoldi\IEEEauthorrefmark{1}, \\ Yukai Chen\IEEEauthorrefmark{2}, Alessio Burrello\IEEEauthorrefmark{1}, Enrico Macii\IEEEauthorrefmark{1}, Massimo Poncino\IEEEauthorrefmark{1}, Daniele Jahier Pagliari\IEEEauthorrefmark{1}, Sara Vinco\IEEEauthorrefmark{1}}
\vspace{-0.3cm}
\IEEEauthorblockA{\\\IEEEauthorrefmark{1}Dept. DAUIN, Politecnico di Torino, Italy, name.surname@polito.it\
    }
    \vspace{-0.4cm}
\IEEEauthorblockA{\\\IEEEauthorrefmark{2} IMEC, Leuven, Belgium, name.surname@imec.be\
    }
}

\maketitle
\copyrightnotice
\begin{abstract}
Virtual Platforms (VPs) enable early software validation of autonomous systems' electronics, reducing costs and time-to-market. While many VPs support both functional and non-functional simulation (e.g., timing, power), they lack the capability of simulating the environment in which the system operates. In contrast, {robotics simulators lack} %
accurate timing and power features. This {twofold shortcoming} limits the effectiveness of the design flow, as the designer can not fully evaluate the features of the solution under development. This paper presents a novel, fully open-source framework bridging this gap by integrating a robotics simulator (Webots) with a VP for RISC-V-based systems (MESSY). {The framework} enables a holistic, mission-level, energy-aware co-simulation of electronics in their surrounding environment, streamlining the exploration of design configurations and advanced power management policies.
\end{abstract}

\vspace{-0.2cm}

\vspace{-0.2cm}
\section{Introduction}
\vspace{-0.1cm}
Virtual Platforms (VPs) enable comprehensive system modeling and simulation before physical production~\cite{VP_Bartolini} and are thus a crucial resource in the design of modern embedded systems, characterized by heterogeneity and tight integration with the physical environment.

State-of-the-art VPs cover system functionality together with non-functional aspects such as timing, power consumption, or thermal behavior~\cite{VP_Lora, messy,9969885}. However, their focus is {mostly on software development, thus their scope is restricted to digital System-on-Chips (SoC), together with their}
auxiliary electronic components (sensors, actuators, batteries, etc.)~\cite{VP_Lora, sara_multilayer, messy}. In this perspective, the surrounding environment is normally considered a secondary aspect, either neglected or left as a ``user-provided'' input. Nonetheless, for robots and autonomous systems, evaluating the optimization of functional and non-functional aspects (e.g., power management policies) in relation to the environment in which they operate is fundamental. For example, the battery lifetime of an Unmanned Aerial Vehicle (UAV) can not be optimized without considering mission parameters such as cruise speed and elevation, its weight, and the accuracy of its autonomous navigation software~\cite{lamberti_object_detection} which strongly depend on the surrounding environment and the scope of the mission.
On the other hand, robotics simulators~\cite{Webots, gazebo,nvidia-isaac-sim} are insufficient to address this problem. They are capable of modeling complete 3D environments, as well as robots' sensors, actuators, and digital computations, but they are restricted to a high-level behavioral simulation of the electronics subsystem, thus neglecting precise modeling of the timing or power/energy consumption of the digital subsystem.

In this work, we propose to overcome this twofold shortcoming by proposing \textbf{\mebots{}}, a tool that connects a comprehensive digital VP capable of functional and non-functional modeling with a full-fledged robotic simulator~\cite{Webots}.
The choice of VP fell on \textbf{ME}SSY \cite{messy}, which extends the RISC-V Instruction Set Simulator (ISS) GVSoC \cite{gvsoc} with the SystemC-AMS simulation of external components (such as virtual sensors/actuators) and power-related devices (batteries, harvesters, DC/DC converters, etc.) \cite{systemcams}. For the robotics simulator, we chose We\textbf{bots}~\cite{Webots}, a popular software capable of modeling dynamic objects such as robots and drones in complex 3D environments. Their integration in a holistic co-simulation framework allows to combine detailed timing and power estimation with the simulation of flight dynamics as an effect of mechanical and environmental factors.

To prove the effectiveness of the proposed solution, we applied \mebots{} to the simulation of a Crazyflie nano-UAV~\cite{crazyflie} in the context of a gate-traversal application~\cite{stargate}, validating and optimizing the power/energy behaviour of the system. This analysis proves that {\mebots{} enables: 1) a \textit{Design Space Exploration (DSE) process} aimed at selecting HW and SW configurations that optimize specific non-functional metrics (e.g., battery lifetime) before the creation of a physical prototype; 2) the \textit{study and optimization of context-dependent power management policies}, that require both the availability of detailed information on the power state of the system and inputs from the external environment.

The rest of the paper is organized as follows: Section II provides essential background and overviews of related works; Section III details our methodology; Section IV presents the experimental results, and Section V concludes the paper. The complete code of \mebots{} can be found at \href{https://github.com/eml-eda/messy}{https://github.com/eml-eda/messy} \cite{mebots}.

\vspace{-0.1cm}
\section{Related Works}
\label{sec:background}
\vspace{-0.1cm}

\subsection{SoC Virtual Platforms}
\vspace{-0.1cm}
VPs of digital SoCs have been developed for decades~\cite{sim-survey} and differ by level of accuracy and focus.

\begin{table}
    \centering
    \caption{Related works: VPs (top) and Robotic Simulators (bottom)}
    \label{tab:related}
    \vspace{-0.2cm}
    \resizebox{\columnwidth}{!}{\begin{tabular}{|c|c|c|c|c|c|}
    \cline{2-6}
    \multicolumn{1}{c|}{} & RISC-V/ARM & Off-chip & Power & Environment & License \\
    \multicolumn{1}{c|}{} & support & components & modeling & \& flight &  \\
    \hline
    QEMU & \yes & \no & \no & \no & GPLv2 \\
    \hline
    Renode & \yes & \yes & \no & \no & MIT \\
    \hline
    gem5 & \yes & \no & $\approx$ & \no & BSD \\
    \hline
    Spike & \yes & \no & \no & \no & BSD-3 \\
    \hline    
    RISC-V-TLM & \yes & \no & \no & \no & GPLv3 \\
    \hline
    GVSoC & \yes & \no & \yes & \no & Apache 2.0 \\
    \hline
    MESSY & \yes & \yes & \yes & \no & Apache 2.0 \\
    \hline\hline
    Webots & \no & \yes & $\approx$ & \yes & Apache 2.0 \\
    \hline
    Gazebo & \no & \yes & $\approx$ & \yes & Apache 2.0 \\
    \hline
    IsaacSim & $\approx$ (HIL) & \yes & \no & \yes & NVIDIA EULA \\
    \hline
    AirSim & $\approx$ (HIL) & \yes & \no & \yes & MIT \\
    \hline\hline
    \rowcolor{LightCyan}
    MEbots & \yes & \yes & \yes & \yes & Apache 2.0 \\
    \hline
    \end{tabular}}  
    \vspace{-0.2cm}
\end{table}

QEMU \cite{qemu,qemu_paper} is an open-source, versatile emulator and virtualizer of a target architecture that supports a wide range of instruction sets (e.g., ARM, RISC-V) with a focus on functional simulation. %
Renode \cite{renode,engproc2024079052} is designed to simulate complete electronic systems with RISC-V support;
{besides functionality, it supports basic timing modeling, yet with low accuracy (minimum resolution of $1\mu s$ and CPU performance modelled only with the average Million of Instructions per Second - MIPS), but lacks power/energy (and external environment) modeling entirely.}
Gem5 \cite{gem5,6322869} is highly regarded for its flexibility and comprehensive support for multiple ISAs, including RISC-V. It offers a variety of CPU models and supports detailed hardware modeling. While supporting accurate timing simulation and state- or equation-based power models, those are limited only to digital SoC modules, without consideration of external components such as batteries and actuators (e.g. motors).

In addition to these multi-platform solutions, multiple VPs have been recently developed specifically for the RISC-V ecosystem. Spike~\cite{spike,9180589} is primarily designed for functional simulation and validation without providing performance assessments, nor any kind of power/energy estimation. RISC-V-TLM~\cite{riscv-tlm,riscv_tlm} %
provides a simple and extensible simulation that abstracts communication details and includes only essential peripherals, enabling early virtual prototyping. However, its simplicity also comes with limitations, as it may not fully capture the complexity of real-world hardware interactions. GVSoC \cite{gvsoc} is an open-source event-driven VP specifically designed for PULP systems~\cite{pulp}, with accurate timing modeling and support for multi-core clusters, multi-level memory hierarchies, and various I/O peripherals and accelerators. LUT-based power models can be defined in GVSoC, but they are limited to SoC components only, meaning the power behavior of external devices cannot be modeled.

MESSY~\cite{messy} is a recently proposed open-source framework that wraps a timing-accurate VP (currently GVSoC) into a top-level system description (in SystemC-AMS) to support the modeling of power behavior of each component (including power consumption, generation, conversion, and storage). However, MESSY does not support the simulation of environmental or mechanical aspects of the system. 
\vspace{-0.1cm}
\subsection{Robotic Simulators}
3D world simulators are essential for the development of complex autonomous systems, as they enable the validation and optimization of software components that determine how the system engages with its surroundings, e.g., training and testing of machine learning models for perception, the validation of control policies for navigation, obstacle avoidance, etc~\cite{stargate}.
Webots is a popular simulator, offering accurate physics modeling capabilities for robots and environmental objects, as well as a vast library of pre-defined components and sensors/actuators (e.g., proximity sensors, cameras) \cite{Webots,robotics12020053}. 
Gazebo \cite{gazebo,robotics12020053} is widely regarded as the most popular open-source alternative to Webots, and it allows to choose from different physics engines for simulation. 
IsaacSim~\cite{nvidia-isaac-sim,10374780} and AirSim \cite{airsim,pmlr-v123-madaan20a} allow for detailed particle physics, to closely mirror real-world dynamics.

While all these tools allow the definition of custom controllers for the robots, {in most of them} the simulation is entirely behavioral, i.e., it completely neglects the detailed architecture of the HW on which the controller SW will be eventually deployed. This implies that: i) in most cases, it is not possible to validate the \textit{actual code} that will be deployed on the robot; ii) the controller reacts instantaneously, or its timing is modeled at a very high level (i.e., with a fixed, user-defined execution delay); iii) {other non-functional properties, in particular power/energy behavior, either are completely ignored (e.g., IsaacSim and AirSim) or can be described with some limitations (e.g., Gazebo imposes limitations by supporting only a single battery per model \cite{gazebo-battery}, while Webots restricts power consumption modeling to motor components \cite{Webots-battery}).}
Notably, IsaacSim and AirSim provide some degree of hardware integration through hardware-in-the-loop (HIL) simulations, allowing seamless interaction with physical systems for more comprehensive testing. However, their hardware support remains limited (e.g, IsaacSim primarily targets the Jetson series~\cite{nvidia-jetson}) and does not account for alternative architectures such as the RISC-V based ones.

In this work, we chose Webots as a starting point to benefit from its wide library of pre-defined components, and its good balance between ease of use, simulation fidelity and resource efficiency. However, our methodology is, for the most part, orthogonal to the selected world simulator and can be extended to any simulator previously described.

\vspace{-0.2cm}
\section{Methodology}
\vspace{-0.2cm}

\begin{figure*}
    \centering
    \includegraphics[width=0.9\linewidth]{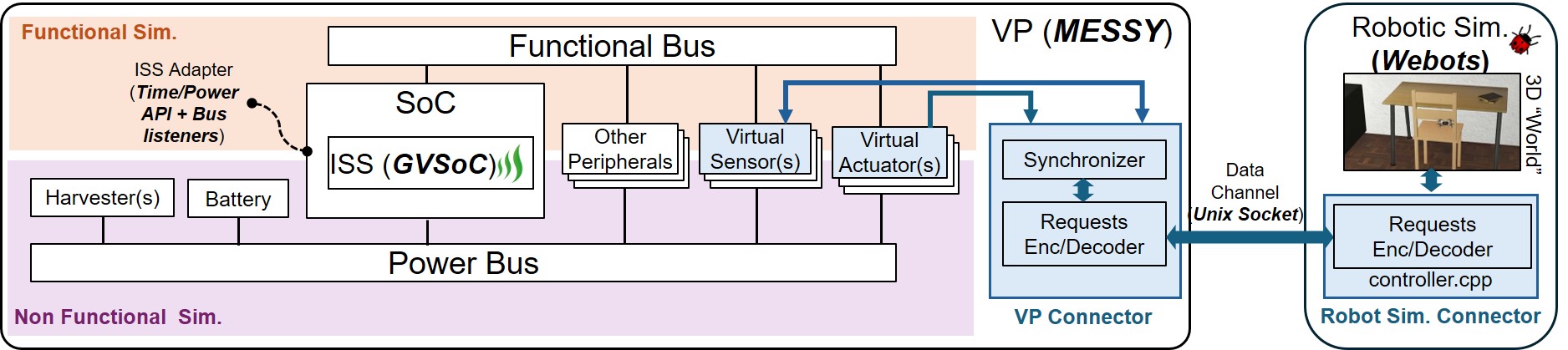}
    \vspace{-0.2cm}
    \caption{\mebots{} architecture: the blue boxes highlight the components involved in the connection between a HW-SW VP (MESSY with the GVSoC ISS, left) with a robotic simulator (Webots, right).}
    \label{fig:framework-architecture}
    \vspace{-0.4cm}
\end{figure*}

\subsection{General Framework Architecture}
\label{sec:arch}
\vspace{-0.1cm}

Fig.~\ref{fig:framework-architecture} shows a high-level architectural diagram of our simulation framework, which connects a full-system VP (MESSY) with a robotic simulator (Webots). The figure highlights in blue the components involved in the connection.

The proposed methodology is orthogonal to the specific tools: any VP capable of modeling virtual sensors and actuators (i.e., components that interact with the external ``environment'') can be instrumented as described in the following, and %
Webots could be replaced by one of the alternatives in Table \ref{tab:related}. This section sketches the main idea behind the construction of \mebots, while details related to the choice of %
MESSY and Webots
will be discussed in the next section. 

\paragraph{Rationale} The link between any SoC VP and a robotic simulator is at the level of sensors and actuators: the sensors must indeed collect data from the environment (e.g., images) to provide input to the computing units, and the actuators must put into practice the {control} strategies elaborated by SW.  
This is where the co-simulation can be built by adding a \textit{Connector}, in charge of implementing data exchange and time synchronization between the two frameworks. 

\paragraph{Role of the connector}
The role of the Connector is twofold, as it is in charge of implementing both data exchange and time synchronization. 
On the VP side, the Connector interacts with a subset of the virtual sensors and actuators. Typical examples are cameras, depth sensors (that receive data coming from the world modeled by the robot simulator), or motor controllers (that influence the robot's movements in said world). Whenever the virtual sensors/actuators receive commands from software and must interact with the environment, their request is %
forwarded to the Connector, that i) \textit{synchronizes} time between the simulators before ii) \textit{encoding and sending} the request over a data channel.
On the robot simulator side, the Connector decodes such incoming request, and performs the appropriate action on the environment (e.g., capturing a picture, changing the motors settings). If the request involves a return value (e.g., the picture), the latter is sent back over an inverse path, after an additional synchronization step. 
\paragraph{Time synchronization}
Time synchronization is crucial to ensure that the two simulations proceed in lock-step. Robot simulators usually implement a fixed-timestep physics engine, with a coarse temporal granularity (in the order of ms). %
The implementation of the \textit{Synchronizer} depends on the selected VP's timing-awareness, which is a design choice influenced by use-case needs in terms of simulation accuracy vs speed. For example, a purely functional VP can be synchronized by delaying each of its requests to the robot simulator by one time-step (i.e., the minimum non-instantaneous delay possible in this scenario), while for a more timing-accurate VP, the time associated with each virtual sensor request can be aligned to the next closest robot simulator timestep. %
Since GVSoC supports close to cycle-accurate execution, our implementation follows the second approach, as we advance the robot simulator by an amount of time that matches the software's progress, rounding to the nearest simulator timestep.

\vspace{-0.2cm}
\subsection{\mebots{} Implementation: MESSY and Webots}
\label{sec:impl}

\subsubsection{MESSY for HW-SW virtual prototyping}
MESSY is primarily implemented in SystemC/SystemC-AMS, and focuses on modelling the power and energy behaviour of a complete embedded system, including a SoC, external peripherals, batteries, harvesters and power converters. A simplified scheme of the MESSY architecture is shown on the left of Fig.~\ref{fig:framework-architecture}. MESSY simulation relies on a \textit{functional bus}, implementing an AXI-like protocol handling functionality (i.e., exchange of data) and timing, and a \textit{power bus}, modeling power flows between components through voltage and current ports, e.g., managing the distribution of power from a battery (or a harvester) to multiple loads (SoC, peripherals, etc) through converters.

MESSY is used to instantiate functional and power models of the system components, that can be either sourced from the tool's library or defined ad hoc. It {also} allows to {integrate} an external ISS through a customizable \textit{Adapter}, to achieve accurate HW-SW prototyping. The current version supports GVSoC~\cite{gvsoc}, an ISS for SoCs of the PULP family, popular in ultra-low-power robotics applications~\cite{lamberti_object_detection}. Through the Adapter, GVSoC uses its APIs to advance the SoC simulation in time as needed and to forward the corresponding SoC consumption to the power bus. Moreover, I/O requests to external devices initiated by the PULP SoC are intercepted by a set of listeners, and forwarded to MESSY's virtual peripheral models. We refer readers to \cite{messy} for details. %

\subsubsection{Webots for drone simulation}
Together with the environment and the robotic platform, Webots allows the definition of custom robot \textit{controllers}, managing exchanges of sensors and actuators data and robot operation. 
On the Webots side, the primary required changes are focused on enabling the interception of commands from MESSY and executing corresponding actions. Thus, these changes involve modifying the \texttt{controller.cpp} file. 
Specifically, we implement a controller that {waits} until it receives a new command via the communication channel. When that happens, the command is parsed to extract the necessary arguments and parameters. The associated logic is then executed to perform the specified action (e.g. take a picture and forward it to MESSY). This approach leverages an opcode-like structure, where each command is linked to a specific action, enabling precise control over the Virtual World and its objects through MESSY, while also allowing simple extensibility by just adding a new opcode.

\subsubsection{Connector implementation}
To enable the communication between the two tools, we implemented the above mentioned connector using UNIX sockets. To ensure standardized communication, we chose to transmit JSON-like packets using a C++ library \cite{json-library}. While this approach slightly increases communication overhead, it provides a structured and easily extensible format. Each packet includes the command opcode and, when necessary, an accompanying payload.

In this setup, MESSY operates as the socket client, while Webots serves as the socket server. This architecture is designed to support future scalability by enabling multiple MESSY instances to connect to Webots. Such a setup reflects real-world scenarios where multiple embedded systems, each with its own structure and power model, collaborate to control a single (or multiple) robotic systems. %

Figure \ref{fig:messy_webots_communication} illustrates an example of the communication process, retrieving data from a virtual camera in the 3D environment and adjusting motor speeds based on the output of a neural network running in GVSoC. At the start of the simulation, Webots initializes a socket and 
MESSY connects to it as client. Then, the simulation proceeds in sync.

The flow begins with GVSoC requesting an image from the Camera Module \circled{1}. This request, transmitted over MESSY's functional bus, prompts the Camera Module to generate a \texttt{GET\_DATA} opcode and send it to the Encoder/Decoder (\circled{2}). This automatically triggers a synchronization action, which advances the Webots simulation (if needed), until the first timestep greater than the current MESSY time.
The Encoder then processes the command, formats it into a JSON-compliant packet, and transmits it through the UNIX socket (\circled{3}).
On the Webots side, the incoming packet is intercepted, and the corresponding action is executed (\circled{4}). In this case, the 3D World captures an image, encapsulates it into a new JSON packet, and transmits it back to MESSY. The returned data payload is directed to the module that initiated the request (\circled{5}). In this example, the ISS is notified of the image's readiness by polling a status register (\circled{6}), although interrupt-based communication is also supported.
The ISS then simulates the neural network execution on the received image (\circled{7}), completing the inference task. For subsequent operations (i.e. setting the speed of the motors), the workflow repeats with a different opcode, following a similar sequence (\circled{8}, \circled{9}). In \circled{9}, similarly to \circled{2} (and in any interaction between the two simulators, all of which go through the connector), timesteps are re-synchronized.

This modular and consistent communication pattern ensures smooth integration and synchronization between MESSY and Webots, regardless of the operation being performed.

A key feature of MEbots is its scalable opcode structure, designed for efficient expansion and maintainability. The whole simulation is controlled by a configuration file (also in JSON format) that not only defines the system architecture, including details about each module's sensors, actuators, and power consumption, but also specifies the unique opcodes assigned to each module. Using this configuration file, a code generation mechanism automatically produces a structured code skeleton already implementing the MESSY/Webots connection. Developers only need to extend this skeleton by implementing the necessary logic for each module.
\vspace{-0.1cm}
\begin{figure}
    \centering
    \includegraphics[width=0.9\linewidth]{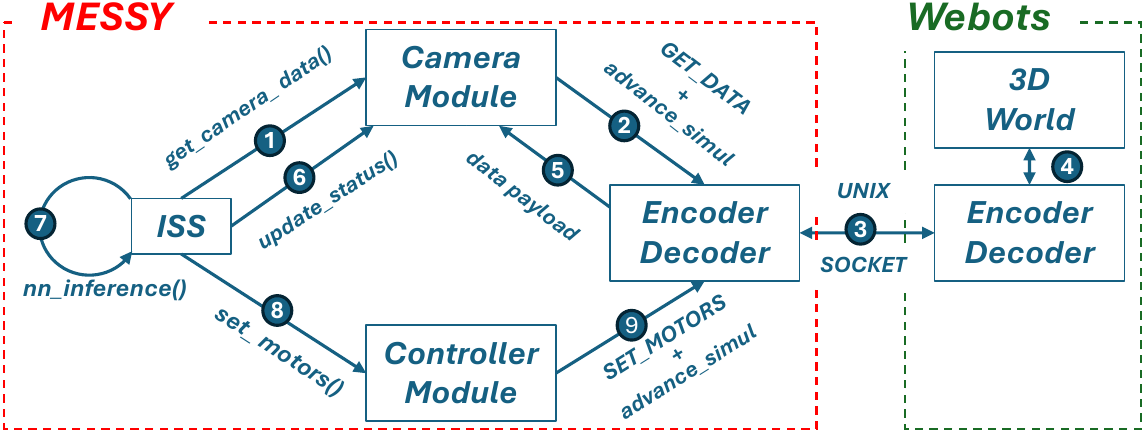}
    \vspace{-0.3cm}
    \caption{Example of MESSY and Webots exchanging messages.}
    \vspace{-0.5cm}
    \label{fig:messy_webots_communication}
\end{figure}

\section{Case Study: Drone Gate Traversal}
\label{sec:usecase}

To validate the proposed methodology, we configured \mebots{} to reproduce a Crazyflie nano-UAV in a gate-traversing scenario, described in \cite{stargate}. 
\vspace{-0.2cm}
\subsection{Drone modeling}
The drone has been modeled in both its functional and power aspects within MESSY, accurately modelling the camera, the motors, the battery and the SoC.
As camera, we selected a Himax HM01B0 integrated into the drone's \textit{AI Deck} \cite{ai-deck}, that captures monochrome images at a resolution of 320x320 pixels. Within MESSY, the camera is implemented as a virtual peripheral with a dedicated contiguous memory area capable of storing the full 320x320-pixel image. Datasheet specifications were used to derive current consumption in the active state (1.75 mA operating at 60 Frames Per Second, FPS) and idle state (0.14 mA, typical standby current under worst-case conditions). {In this case study, GVSoC is used to simulate the GAP8 processor \cite{gap8} on board of the AI Deck. %

{Motors} plays a critical role in power consumption analysis, as they account for nearly 95\% of the total power usage \cite{stargate}. To achieve an accurate representation, we developed a custom power model that considers two key components. $P_{hover}$ is the power required to maintain the drone in a hovering state: it factors in the mass $m$ of the complete system (including the drone body, AI deck, and battery) and it incorporates several datasheet parameters such as the propeller pitch, revolutions per second (rps), and air density. 
$P_{propel}$ instead focuses on the power required for horizontal movement at a steady cruising speed, and it takes into account the total system mass $m$, the propulsion system's efficiency $\eta$, and the cruising speed $v$. By combining these components, the model provides a simple but realistic estimate of the drone's power consumption. %

We selected four different batteries to conduct a comprehensive exploration of battery capacity and weight impact on drone performance: the standard Bitcraze battery (250mAh, 7.1g), a UFX battery (250mAh, 5.8g) \cite{battery_ufx_250}, a Cyclone battery (300mAh, 8.1g) \cite{cyclone_300mah}, and a Lipol battery (350mAh, 9.2g) \cite{lipolbattery_15c_lipo}. These batteries range from a slightly lighter alternative to the standard Bitcraze battery to a larger battery offering 40\% more capacity but at a significant weight increase. %
Batteries have been modeled by extrapolating information from datasheets to build their circuit equivalent models. 
Additionally, we incorporated %
the necessary DC/DC converters.

\vspace{-0.2cm}
\subsection{Gate traversal software}
We built two distinct Webots simulation environments: an easy one, featuring a gate positioned almost directly in front of the drone, %
and a more challenging one, where the gate is near the edge of the drone’s camera Field of View (FoV). This significantly increases the mission's difficulty, as the drone must rely on precise speed control and well-tailored logic to successfully navigate through the gate.
The drone is 9.2cm$\times$9.2cm$\times$2.9cm, and each gate is a square with an internal opening of 40cm$\times$40 cm. The gate frame has outer dimensions of 60cm$\times$60cm, providing a 20cm wide and 3cm thick border. The gates are at a fixed height of 83cm above the ground, ensuring consistency across tests.

{To control the drone, we employed a Convolutional Neural Network (CNN), that processes grayscale camera images downsampled to 168×168 pixels. The CNN architecture is taken from \cite{stargate}, modified to only use camera images and no Time of Flight (ToF). It consists of three sequential blocks, each containing Convolutional Layers, Batch Normalization, and ReLU6 activation. The number of channels increases progressively from 1 to 32. A final Fully Connected layer outputs the predicted yaw rate.} We trained it using the open-source dataset \cite{dataset} following the established data pre-processing used in \cite{stargate}. {We first trained for 100 epochs with a batch size of 256. The chosen optimizer is Adam \cite{adam}, with an initial learning rate of 0.001 and a weight decay of 0.0001.} Subsequently, we applied quantization-aware fine-tuning for 50 epochs using the open-source PLiNIO library, reducing both weights and activations to int8 precision \cite{10272045}. {Finally, the model has been deployed to GVSoC using the MATCH compiler \cite{match}.}

In all experiments, the drone started from a predefined position, ascended to an altitude of 1 meter, and then cruised toward the direction indicated by the CNN at a specific speed, set to $v_{easy} \in \{0.5km/h, 1 km/h, 1.5 km/h\}$ in the easy scenario, and to $v_{hard} \in \{0.1km/h, 0.2 km/h, 0.3 km/h, adaptive\}$ in the hard scenario (as will be elaborated later on).
After successfully passing through the gate, or when the battery SoC reaches a predefined threshold (10\%), the drone lands. 

\vspace{-0.1cm}
\section{Experimental Results}
\vspace{-0.1cm}
\subsection{Battery choice exploration}

The first batch of experiments focused on the simple gate-traversing scenario to evaluate the effectiveness of different batteries. 
Each speed scenario was tested with the four different batteries, and the %
results of these experiments are illustrated in Figure \ref{fig:batteries_easy}, where the x-axis represents the cruising speed of the drone, and the y-axis indicates the {consumed state of charge} (SoC) at the end of the simulation. %

We first isolated the impact of sole battery capacity, by modeling all batteries assuming they had the same weight as the heaviest one. As expected, the smallest-capacity battery consumed the highest amount of charge, and the consumed charge showed a clear proportionality to the battery capacity, as reflected by the total bar heights (fully colored part plus shadowed part).
For instance, the Cyclone battery has a capacity approximately $\sim 15\%$ smaller than the Lipol battery, and this is reflected in a  higher SoC usage, falling within the range of $[16-17]\%$. The slight mismatch w.r.t. the expected proportional difference is due to non-linear battery dynamics and secondary effects (parasitic discharges, aging, etc).

When accounting for the actual weight of each battery (full-color bars), the results show a reduction in the consumed SoC, even if the trend is not modified (a lower total weight implies a lower $P_{hover}$ and $P_{propel}$ and thus a lower $\Delta$SoC). A noteworthy comparison arises between the original battery and the UFX, both of which have identical capacity (250 mAh) but different weight (the original battery is 1.3g heavier compared to the UFX). Across all tested speeds, the latter consistently consumed less SoC, with savings of $3\%$, $3.5\%$, and $2.8\%$ at $0.5 km/h$, $1km/h$, and $1.5km/h$, respectively.  
These savings translate into a potential increase in flight time. To assess this increase, we first validated Bitcraze’s claimed 7-minutes flight duration using our framework with the original battery, obtaining a flight time of around 6 minutes and 50 seconds. With this baseline, the UFX battery could extend the flight time by up to 15 seconds (4\%), thanks to the sole weight reduction. Alternatively, by allowing for a slight increase in the battery pack's weight, we can achieve a significant $33\%$ increase in battery life, at the price of a small reduction in the maximum recommended payload weight. This translates into an additional flight time of around 2 minutes and 30 seconds, which at a forward cruising speed of $1.5km/h$ translates into a covered distance of 62m. Overall, this experiment shows the importance of modelling flight dynamics (e.g. dependence on weight) to precisely estimate power/energy evolution in these systems, as enabled by MEbots.

\begin{figure}
    \vspace{-0.2cm}
    \centering
    \includegraphics[width=0.9\linewidth]{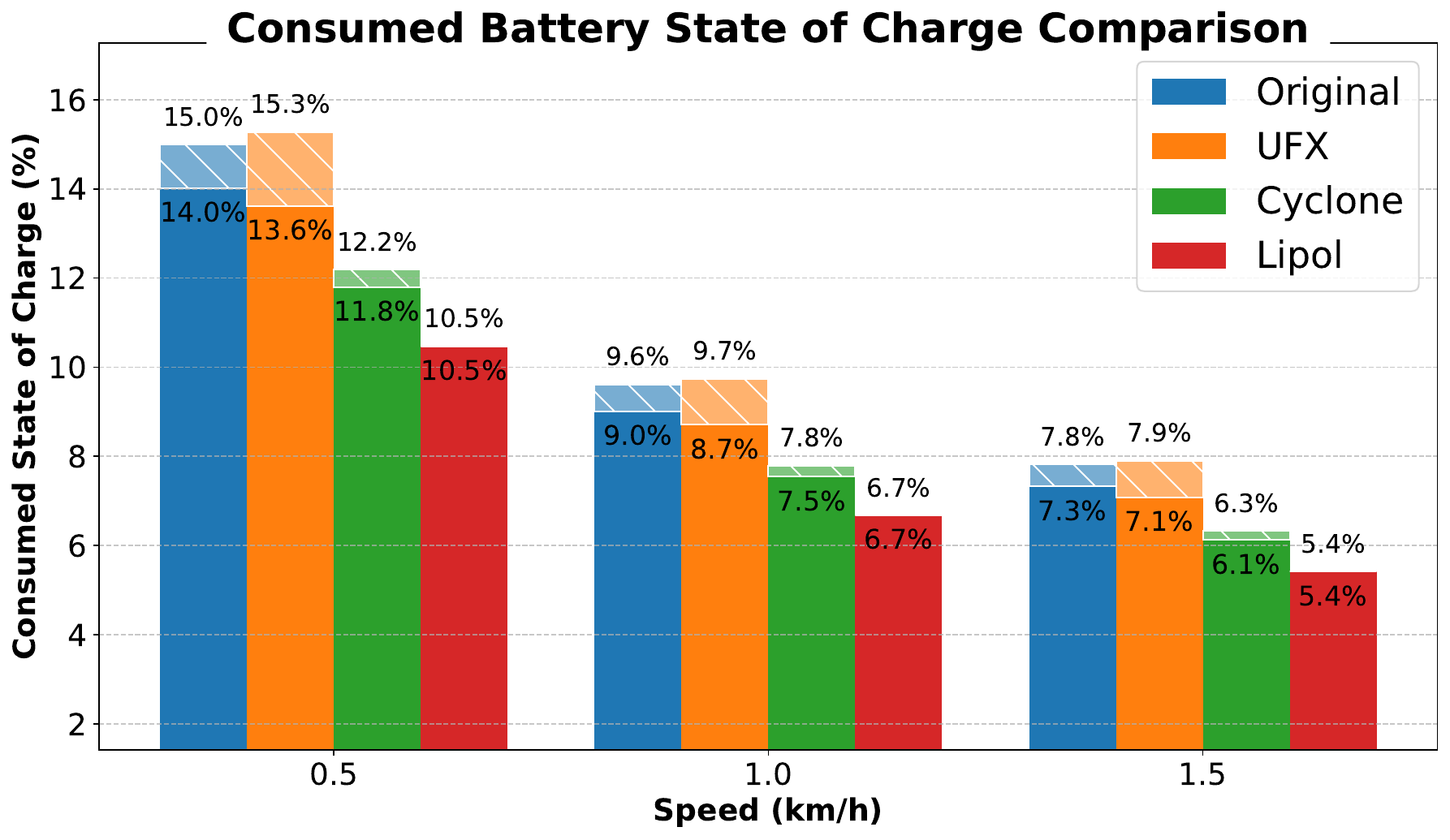}
    \vspace{-0.4cm}
    \caption{Consumed State of Charge at different cruising speed and with different batteries when considering actual battery weight (full color) or the weight of the heaviest battery (shadows), to isolate the impact of capacity.}
    \vspace{-0.3cm}
    \label{fig:batteries_easy}
    \vspace{-0.3cm}
\end{figure}

\begin{figure*}
    \vspace{-0.2cm}
    \centering
    \includegraphics[width=0.9\linewidth]{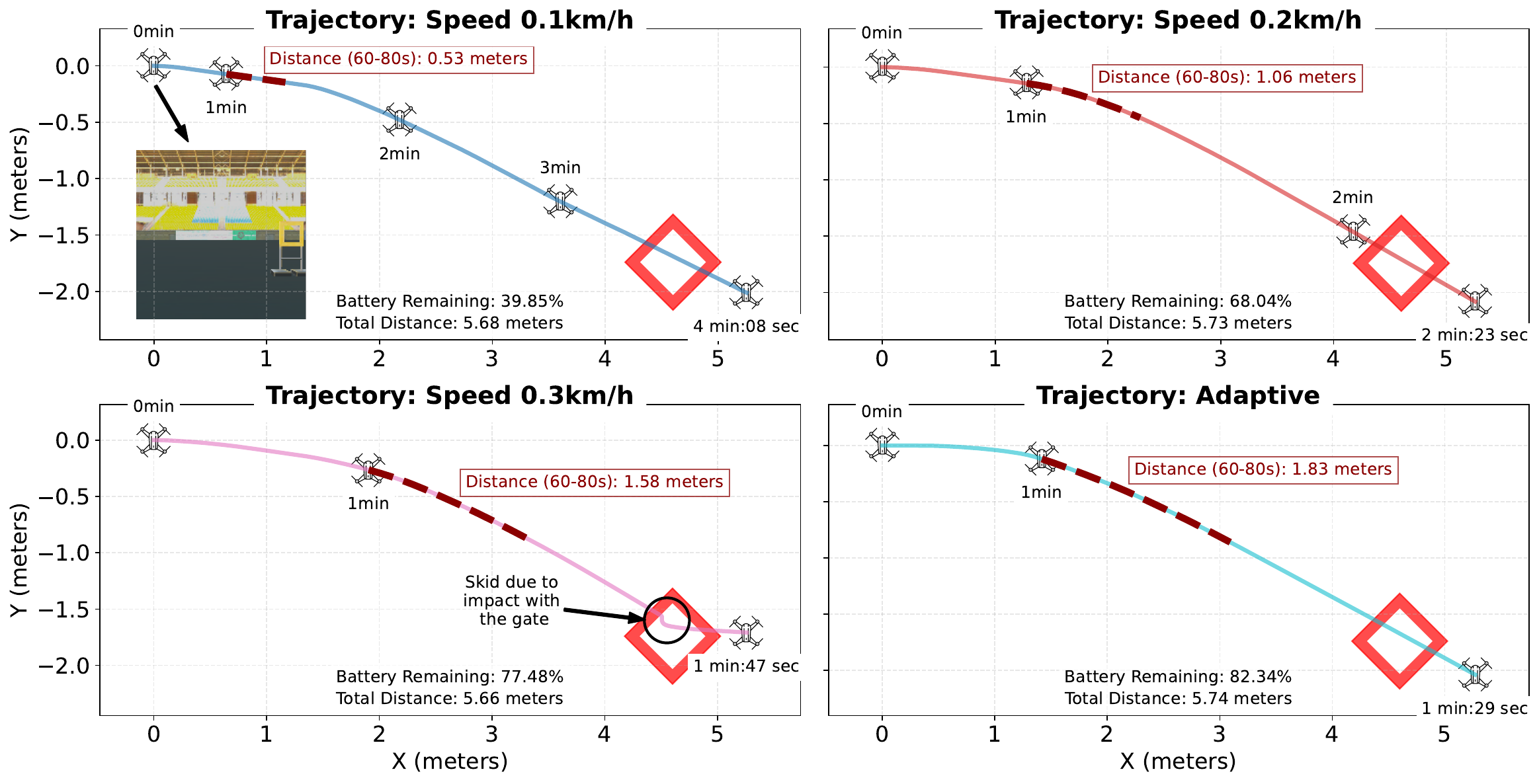}
    \vspace{-0.4cm}
    \caption{Trajectory comparison with standard Bitcraze’s battery at different fixed and adaptive speeds.}
    \vspace{-0.5cm}
    \label{fig:trajectories}
\end{figure*}

\vspace{-0.2cm}
\subsection{Drone Gate-Traversing speed exploration}
In drone racing, a key objective is to pass through gates in the shortest possible time, as %
shorter traversal times correlate with reduced battery consumption, as less energy is required to complete the objective. 
Using the challenging environment described above, we demonstrate how \mebots{} can be used to evaluate different flying strategies in terms of total battery consumed and task accomplishment. Notably, relying solely on Webots would not allow for an accurate estimation of total energy consumption, as the traversal time also depends on the accuracy of the CNN's prediction and on its reaction latency (both of which are precisely simulated in the VP).

The initial view of the drone is shown in the top-left subplot of Figure \ref{fig:trajectories}. The figure presents four subplots, each comparing different flying strategies: three with constant cruising speeds of $0.1 km/h$, $0.2 km/h$, $0.3 km/h$, and one employing an $adaptive$ speed control, all with the default battery.

At constant speeds of $0.1 km/h$ and $0.2 km/h$, the drone successfully enters the gate without issues. However, the time required to reach the gate is substantial, leading to lower remaining battery life at the simulation's end. The total time for $0.2 km/h$ is not exactly half that of $0.1 km/h$ due to the additional time required for takeoff, landing, and acceleration to cruising speed. To confirm that cruising speeds were correctly implemented, we measured the distance covered by the drone during a portion of the path at quasi-cruising speed; the distance covered in 20 seconds at $v = 0.2km/h \equiv 0.056m/s$ is almost 1.1 meters, double that at $ v = 0.1 km/h$.

Among the constant speeds, $0.3 km/h$ is the fastest while still enabling quasi-successful gate entry. As visible in the trajectory, the drone exhibits a minor skid due to an impact on the gate structure when traversing it, likely caused by the higher speed. Despite this, the drone re-stabilizes and traverses successfully. This establishes $0.3 km/h$ as the upper limit for constant cruising speed. For completeness, we also tested higher speeds of $0.4 km/h$ and $0.5 km/h$, but in these cases, the drone failed to enter the gate, and we omit the corresponding plots for clarity. 

The adaptive flying strategy, instead, implements a
simple real-time tuning mechanism of the drone's speed based
on its alignment with the gate. Specifically, the yaw rate predicted by the CNN is a value in $[-1,1]$, and can be used to estimate how centered is the gate within the drone's field of view (lower absolute values correspond to a more centered gate). Thus, our adaptive policy averages the last 10 CNN predictions, and sets the drone speed to a constant 1 km/h when $\|\mathrm{yaw}_{avg}\| \le 0.3$, while decreasing it by 0.05 km/h, when  $\|\mathrm{yaw}_{avg}\| > 0.3$.
Using this adaptive technique, the required time to reach the gate was reduced to 1 minute and 29 seconds, representing improvements of $64\%$, $38\%$, and $17\%$ compared to constant speeds of $0.1 km/h$, $0.2 km/h$, and $0.3 km/h$, respectively. Additionally, the adaptive approach resulted in significant energy savings, with the remaining SoC increasing from $39.85\%$ in the worst-case scenario to $82.34\%$, with an impressive $42.5\%$ improvement. Despite the dynamic speed adjustments, the total distance covered was nearly identical to that of the constant-speed cases, showing only a $1\%$, $0.17\%$, and $1.4\%$ increase compared to $0.1 km/h$, $0.2 km/h$, and $0.3 km/h$, respectively.

These results clearly highlight the importance of environment modelling when performing DSE to balance performance and energy efficiency. By modeling the entire system, including functional aspects like control logic and extra-functional properties such as energy consumption, MEbots permits the analysis of how different design choices influence the ability to achieve a Pareto-optimal trade-off between energy efficiency, drone speed and mission accomplishment.

\vspace{-0.1cm}
\section{Conclusions and Future Works}
\vspace{-0.1cm}
This paper presented \mebots, a framework that combines a drone simulator with a VP equipped with an ISS and extra-functional support. The application to a nanodrone proved that \mebots{} simplifies early validation of software strategies and exploration of alternative design directions. In such a challenging and multi-disciplinary scenario, \mebots{} is thus a crucial resource that can help designers achieve better optimization. As future work, we would like to investigate \mebots{} support for other robotics scenarios, and further explore the software optimization that it enables.

\clearpage
\bibliographystyle{IEEEtran}
\bibliography{bibliography} 

\vspace{12pt}

\end{document}